\documentclass[10pt,twocolumn,letterpaper]{article}

\usepackage{cvpr}
\usepackage{times}
\usepackage{epsfig}
\usepackage{graphicx}
\usepackage{amsmath}
\usepackage{amssymb}

\usepackage{graphicx}
\usepackage{amsmath,amssymb} % define this before the line numbering.
\usepackage{color}
\usepackage{booktabs}
\usepackage{mathastext}
\usepackage{multirow}
\usepackage{comment}
\usepackage[normalem]{ulem}

\usepackage[pagebackref=true,breaklinks=true,letterpaper=true,colorlinks,bookmarks=false]{hyperref}

\cvprfinalcopy % *** Uncomment this line for the final submission

\def\cvprPaperID{2705} % *** Enter the CVPR Paper ID here

\ifcvprfinal\fi
\begin{document}

%%%%%%%%% TITLE
\title{Mask Scoring R-CNN}

\author{Zhaojin Huang$^{\dag}$$^*$ \ \  Lichao Huang$^{\ddag}$ \ \  Yongchao Gong$^{\ddag}$ \ \   Chang Huang$^{\ddag}$ \ \  Xinggang Wang$^{\dag}$\\
        $^{\dag}$Institute of AI, School of EIC, Huazhong University of Science and Technology \\
        $^{\ddag}$Horizon Robotics Inc. \\
	{\tt\small \{zhaojinhuang,xgwang\}@hust.edu.cn \{lichao.huang,yongchao.gong,chang.huang\}@horizon.ai}
}

\maketitle
%\thispagestyle{empty}

%%%%%%%%% ABSTRACT
\begin{abstract}

%   Letting a deep network be aware of the quality of its own predictions is an interesting yet important problem. 
%   In the task of instance segmentation, we study this problem and propose Mask Scoring R-CNN which contains a network block to learn the quality of the predicted instance masks. 
%   The mask quality is quantified using the IoU between instance mask and its ground truth. 
%   The proposed network block takes the instance feature and the predicted mask together to regress the mask IoU.
%   The mask scoring strategy helps the instance segmentation network to obtain more reliable detection confidence and brings new information to boost semantic segmentation results.
%   By extensively evaluated on the COCO dataset, Mask Scoring R-CNN consistently and noticeable outperforms the state-of-the-art Mask R-CNN. 
%   The improved results confirm that the proposed mask scoring network is new direction for improving instance segmentation.

  Letting a deep network be aware of the quality of its own predictions is an interesting yet important problem. In the task of instance segmentation, the confidence of instance classification is used as mask quality score in most instance segmentation frameworks. However, the mask quality, quantified as the IoU between the instance mask and its ground truth, is usually not well correlated with classification score. 
  In this paper, we study this problem and propose Mask Scoring R-CNN which contains a network block to learn the quality of the predicted instance masks.  
  The proposed network block takes the instance feature and the corresponding predicted mask together to regress the mask IoU.
  The mask scoring strategy calibrates the misalignment between mask quality and mask score, and improves instance segmentation performance by prioritizing more accurate mask predictions during COCO AP evaluation.  
  By extensive evaluations on the COCO dataset, Mask Scoring R-CNN brings consistent and noticeable gain with different models, and outperforms the state-of-the-art Mask R-CNN. We hope our simple and effective approach will provide a new direction for improving instance segmentation. The source code of our method is available at \url{https://github.com/zjhuang22/maskscoring_rcnn}.
\end{abstract}

\vspace{-10mm}
\let\thefootnote\relax\footnote{$^*$ The work was done when Zhaojin Huang was an intern in Horizon Robotics Inc.}

%%%%%%%%% BODY TEXT
\section{Introduction}

    %Modern Deep Convolution Neural Network (ConvNets) has greatly driven the development of computer vision, especially for classification, object detection, semantic and instance segmentation fields. Object detection and semantic segmentation tasks based on ConvNets could always achieve remarkable results. 
   %------------------------------------------------------Modified
    
    Deep networks are dramatically driving the development of computer vision, leading to a series of state-of-the-art in tasks including classification~\cite{krizhevsky2012imagenet, he2016deepresnet,tang2017deep}, object detection~\cite{girshick2014rich, huang2015densebox, redmon2016youyolo, liu2016ssd, ren2015faster, tang2018object}, semantic segmentation~\cite{long2015fully, chen2018deeplab, zhao2017pyramidpspnet,huang2018ccnet} \etc. 
    From the development of deep learning in computer vision, we can observe that the ability of deep networks is gradually growing 
    from making image-level prediction \cite{krizhevsky2012imagenet} to region/box-level prediction \cite{girshick2014rich}, pixel-level prediction \cite{long2015fully} and instance/mask-level prediction \cite{he2017maskrcnn}. The ability of making fine-grained predictions requires not only more detailed labels but also more delicate network designing. 
    
    In this paper, we focus on the problem of instance segmentation, which is a natural next step of object detection to move from coarse box-level instance recognition to precise pixel-level classification. 
    %Specifically, this work discusses how to score the instance segmentation hypothesis. It is important to score an instance segmentation hypothesis precisely in most applications, since the evaluation metric is made based on the scores. In the most challenging instance segmentation dataset COCO \cite{lin2014microsoftcoco}, the performance of instance segmentation is evaluated using precision-recall curves and mean average precision (mAP). If the instance segmentation hypothesis are not properly scored, they could be regarded as false positive or false negative, resulting a decrease of mAP. 
    Specifically, this work presents a novel method to score the instance segmentation hypotheses, which is quite important for instance segmentation evaluation. The reason lies in that most evaluation metrics are defined according to the hypothesis scores, and more precise scores help to better characterize the model performance. 
    For example, precision-recall curves and average precision (AP) are often used for the challenging instance segmentation dataset COCO \cite{lin2014microsoftcoco}. If one instance segmentation hypothesis is not properly scored, it might be wrongly regarded as false positive or false negative, resulting in a decrease of AP.
    
    % In practical applications, scoring an instance segmentation hypothesis is important, since decisions are made based on the score. In the most challenging instance segmentation benchmark, the COCO dataset \cite{lin2014microsoftcoco}, the performance of instance segmentation is evaluated using precision-recall curves and mean average precision (mAP); if the instance segmentation hypothesis are not properly scored, mAP will be decreased. 
    
    % However, in most instance segmentation pipelines, such as Mask R-CNN and MaskLab \cite{}, the instance mask is predicted using a pixel-wise cross-entropy loss and the score of the instance mask is predicted by a classifier applied on the proposal feature. Scoring instance mask only using a classification is unreasonable. Since the classification score only distinguishes the proposal's semantic category. The classifier does not care about the completeness of the instance mask.
    % When we visualize images in Fig.~\ref{fig:sample}, we can discover that an instance segmentation hypothesis may get a good box-level localization result and a high classification score, but its mask is bad. These samples are harmful to the masks results.
    
\begin{figure*}[!htp]
\centering
\includegraphics[width=1.0\linewidth]{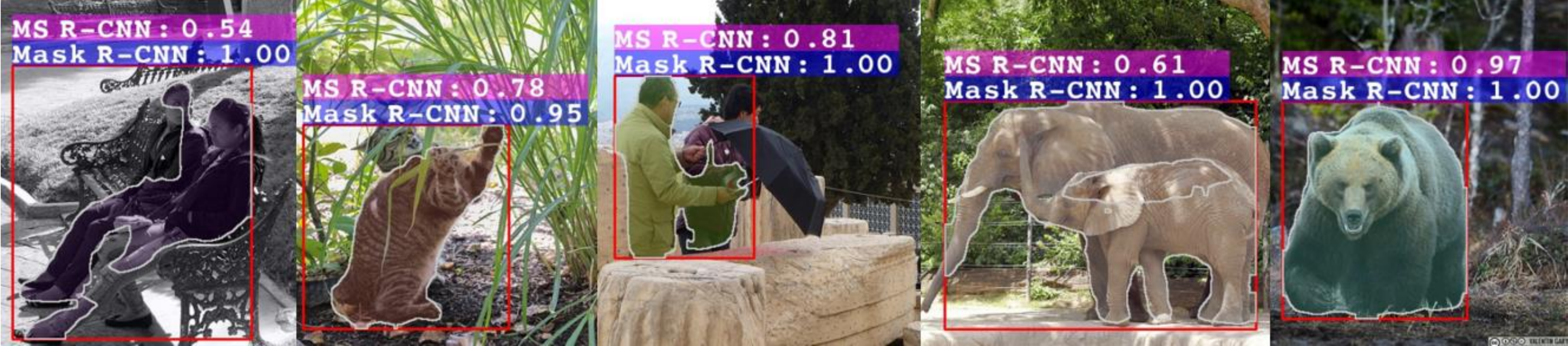}
\caption{Demonstrative cases of instance segmentation in which bounding box has a high overlap with ground truth and a high classification score while the mask is not good enough. The scores predicted by both Mask R-CNN and our proposed MS R-CNN are attached above their corresponding bounding boxes. The left four images show good detection results with high classification scores but low mask quality. Our method aims at solving this problem. The rightmost image shows the case of a good mask with a high classification score. Our method will retrain the high score. As can be seen, scores predicted by our model can better interpret the actual mask quality.}
\label{fig:sample}
\end{figure*}

    However, in most instance segmentation pipelines, such as Mask R-CNN \cite{he2017maskrcnn} and MaskLab \cite{chen2017masklab}, the score of the instance mask is shared with box-level classification confidence, which is predicted by a classifier applied on the proposal feature. 
    %Using classification score as mask confidence is inappropriate since the classification score only distinguishes the proposal's semantic category, and it does not aware of the quality and completeness of the instance mask. 
    It is inappropriate to use classification confidence to measure the mask quality since it only serves for distinguishing the semantic categories of proposals, and is not aware of the actual quality and completeness of the instance mask.
    The misalignment between classification confidence and mask quality is illustrated in Fig.~\ref{fig:sample}, 
    %{in which an instance segmentation hypothesis may get a good box-level localization result and a high classification score, but its mask is bad. These samples are harmful to the masks results.} 
    where instance segmentation hypotheses get accurate box-level localization results and high classification score, but the corresponding masks are inaccurate. Obviously, scoring the masks using such classification score tends to degrade the evaluation results.
    
    Unlike the previous methods that aim to obtain more accurate instance localization or segmentation mask, our method focuses on scoring the masks. To achieve this goal, our model learns a score for each mask instead of using its classification score. For clarity, we call the learned score mask score.
    % As the mAP metric shows, the IoU between detection and ground-truth is used to describe the results. Inspired by this, our method is going to learn an IoU (called MaskIoU) between predicting masks and ground truth masks. 
    
    Inspired by the AP metric of instance segmentation that uses pixel-level Intersection-over-Union (IoU) between the predicted mask and its ground truth mask to describe instance segmentation quality, we propose a network to learn the IoU directly. 
    %In the mAP metric of instance segmentation, pixel-level Intersection-over-Union (IoU) is adopted to describe the mask quality. Therefore, we are inspired to learn the IoU directly using a network so that the network can be used to estimate the mask quality.
    In this paper, this IoU is denoted as MaskIoU. Once we obtain the predicted MaskIoU in testing phase, mask score is reevaluated by multiplying the predicted MaskIoU and classification score. Thus, mask score is aware of both semantic categories and the instance mask completeness.

    % We only need to add a branch (called MaskIoU Head) in the Mask R-CNN backbone. The MaskIoU head is parallel with the R-CNN head and Mask head. For making the predicted MaskIoU is aware of the masks, we need to feed the masks as an input of MaskIoU head. We have conducted extensive experiments based on Mask R-CNN, and the results are effective in many benchmarks. 

    Learning MaskIoU is quite different from proposal classification or mask prediction, as it needs to ``compare" the predicted mask with object feature. Within the Mask R-CNN framework, we implement a MaskIoU prediction network named MaskIoU head. It takes both the output of the mask head and RoI feature as input, and is trained using a simple regression loss. 
    We name the proposed model, namely Mask R-CNN with MaskIoU head, as Mask Scoring R-CNN (MS R-CNN). 
    Extensive experiments with our MS R-CNN have been conducted, and the results demonstrate that our method provides consistent and noticeable performance improvement attributing to the alignment between mask quality and score. 
    
    %Besides Mask R-CNN, the proposed MaskIoU net can be be easily plugged into any other existing instance segmentation networks for improving performance. %{Besides Mask R-CNN, the proposed MaskIoU net can be be easily plugged into any other existing instance segmentation networks for improving performance。 我们并没有把这个模块放到别的instance segmentation networks实验过，这样说感觉不好... by zjhuang}
    
    In summary, the main contributions of this work are highlighted as follows:
    \begin{enumerate}
    \item We present Mask Scoring R-CNN, the first framework that addresses the problem of scoring instance segmentation hypothesis. It explores a new direction for improving the performance of instance segmentation models. By considering the completeness of instance mask, the score of instance mask can be penalized if it has high classification score while the mask is not good enough. 
    %  We present a MaskIoU head to score the instance masks. Then, the score of instance segmentation hypothesis is calculated by multiplying the predicted MaskIoU and the classification score. The score can penalize the instance segmentation hypothesis which have high classification scores while the instance masks are not good enough. 
    \item Our MaskIoU head is very simple and effective. Experimental results on the challenging COCO benchmark show that when using mask score from our MS R-CNN rather than only classification confidence, the AP improves consistently by about $1.5\%$ with various backbone networks.
    % \item  
    % Our method is the first work that addresses the problem of scoring instance segmentation hypothesis, which explorers a new direction for improving the performance of instance segmentation models.
    \end{enumerate}

    %Most of these algorithms addressing on refining final predictions indeed improve the accuracy of masks and bounding boxes. However, for a long period of time, many researchers focus on designing a more powerful RCNN network or applying post-processing means, denseCRF or its variants, to boost their final performance, in an nonessential aspect. We propose an intuition technique, named 'Mask Score', dedicating to ease our network ##maskscore作用##
    
    %Instance segmentation methods are dominated by Mask R-CNN addressing object detection and semantic segmentation also benefits from it. Mask R-CNN firstly detects an object and then uses a mask branch to get the masks of the object. Besides, most algorithms aim to design a better detector or semantic segmentation model to refine masks.

    %{\color{red}{“The detection score only describes how well the predicted boxes match the ground %truth” detection score 都无法描述how well it matches the ground truth. 请看旷视的IOUNet.  
    %}}
    
    %-----------------------------------------------------

\section{Related Work}

\subsection{Instance Segmentation}

    Current instance segmentation methods can be roughly categorized into two classes. One is detection based methods and the other is segmentation based methods. Detection based methods exploit the state-of-the-art detectors, such as Faster R-CNN~\cite{ren2015faster}, R-FCN~\cite{dai2016rfcn}, to get the region of each instance, and then predict the mask for each region.
    Pinheiro \etal \cite{deepmask} proposed DeepMask to segment and classify the center object in a sliding window fashion.
    Dai \etal \cite{dai2016instance-sensitiveFCN} proposed instance-sensitive FCNs to generate the position-sensitive maps and assembled them to obtain the final masks. 
    FCIS~\cite{fcis} takes position-sensitive maps with inside/outside scores to generate the instance segmentation results.
    He \etal~\cite{he2017maskrcnn} proposed Mask R-CNN that is built on the top of Faster R-CNN by adding an instance-level semantic segmentation branch. Based on Mask R-CNN, Chen \etal~\cite{chen2017masklab} proposed MaskLab that used position-sensitive scores to obtain better results. However, an underlying drawback in these methods is that mask quality is only measured by the classification scores, thus resulting in the issues discussed above.
    
\begin{figure*}[!t]
\centering
\includegraphics[width=1.0\linewidth]{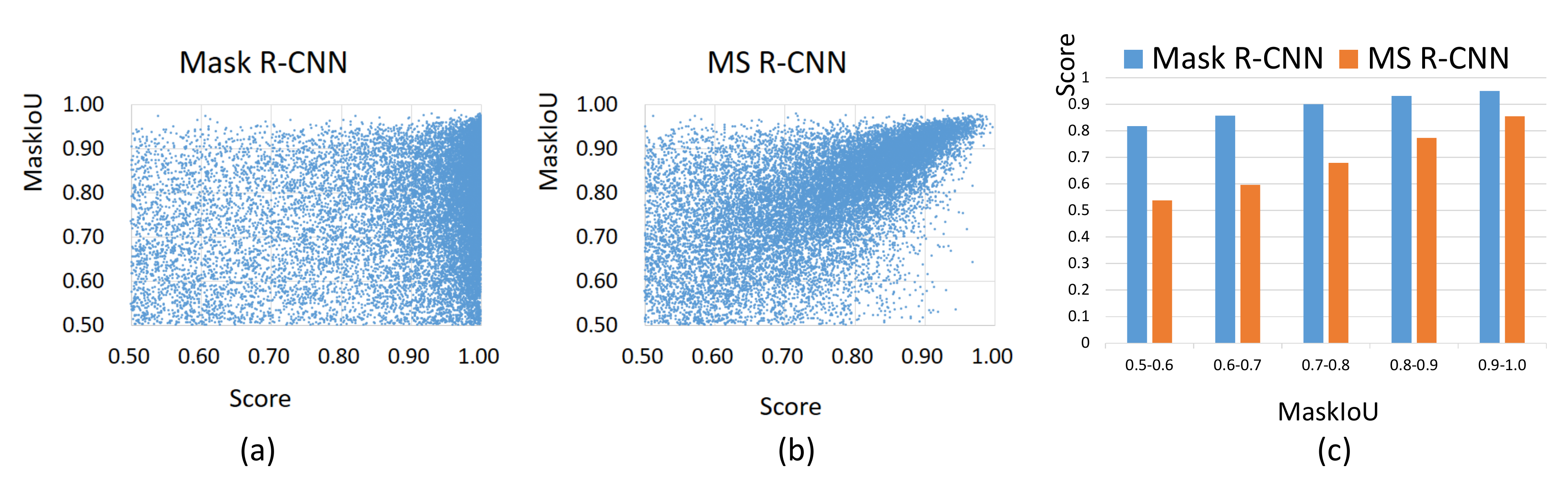}
\caption{Comparisons of Mask R-CNN and our proposed MS R-CNN. (a) shows the results of Mask R-CNN, the mask score has less relationship with MaskIoU. (b) shows the results of MS R-CNN, we penalize the detection with high score and low MaskIoU, and the mask score can correlate with MaskIoU better. (c) shows the quantitative results, where we average the score between each MaskIoU interval, we can see that our method can have a better correspondence between score and MaskIoU.}
\label{fig:comv4}
\end{figure*}         
    
    Segmentation based methods predict the category labels of each pixel first and then group them together to form instance segmentation results. Liang \etal \cite{liang2015proposal} used spectral clustering to cluster the pixels. Other works, such as \cite{jin2016object, kirillov2017instancecut}, add boundary detection information during the clustering procedure. 
    Bai \etal~\cite{bai2017deepwatershed} predicted pixel-level energy values and used watershed algorithms for grouping. 
    Recently, there are some works \cite{newell2017associative, fathi2017semantic, harley2017segmentation, de2017semantic} using metric learning to learn the embedding. Specifically, these methods learn an embedding for each pixel to ensure that pixels from the same instance have similar embedding. Afterwards, clustering is performed on the learned embedding to obtain the final instance labels. 
    As these methods do not have explicit scores to measure the instance mask quality, they have to use the averaged pixel-level classification scores as an alternative.
    
    Both classes of the above methods do not take into consideration the alignment between mask score and mask quality. Due to the unreliability of mask score, a mask hypothesis with higher IoU against ground truth is vulnerable to be ranked with low priority if it has a low mask score. In this case, the final AP is consequently degraded.

\subsection{Detection Score Correction}    

    There are several methods focusing on correcting the classification score for the detection box, which have a similar goal to our method. 
    Tychsen-Smith \etal \cite{tychsen2017improving} proposed Fitness NMS that corrected the detection score using the IoU between the detected bounding boxes and their ground truth. It formulates box IoU prediction as a classification task. Our method differs from this method in that we formulate mask IoU estimation as a regression task.
    Jiang \etal~\cite{jiang2018acquisition} proposed IoU-Net that regressed box IoU directly, and the predicted IoU was used for both NMS and bounding box refinement. 
    In \cite{cheng2018revisiting}, Cheng~\etal discussed the false positive samples and used a separated network for correcting the score of such samples. 
    SoftNMS \cite{bodla2017softnms} uses the overlap between two boxes to correct the low score box.
    Neumann \etal~\cite{relatedsoftmax_fromvgg} proposed Relaxed Softmax to predict temperature scaling factor value in standard softmax for safety-critical pedestrian detection.
    
    Unlike these methods that focus on bounding box level detection, our method is designed for instance segmentation. The instance mask is further processed in our MaskIoU head so that the network can be aware of the completeness of instance mask, and the final mask score can reflect the actual quality of the instance segmentation hypothesis. It is a new direction for improving the performance of instance segmentation.

\section{Method}
\subsection{Motivation}

   In the current Mask R-CNN framework, the score of a detection (\ie, instance segmentation) hypothesis is determined by the largest element in its classification scores. Due to the problems of background clutter, occlusion \etc, it is possible that the classification score is high but the mask quality is low, as the examples shown in Fig.~\ref{fig:sample}. 
   To quantitatively analyze this problem, we compare the vanilla mask score from Mask R-CNN with the actual IoU between the predicted mask and its ground truth mask (MaskIoU).
   Specifically, we conduct experiments using Mask R-CNN with ResNet-18 FPN on COCO 2017 validation dataset. Then we select the detection hypotheses after Soft-NMS with both MaskIoU and classification scores larger than 0.5.
   The distribution of MaskIoU over classification score is shown in Fig.~\ref{fig:comv4}~(a) and the average classification score in each MaskIoU interval is shown in blue in Fig.~\ref{fig:comv4}~(c). These figures show that \textit{classification score and MaskIoU is not well correlated in Mask R-CNN}.
 
    In most instance segmentation evaluation protocols, such as COCO, a detection hypothesis with a low MaskIoU and a high score is harmful. In many practical applications, it is important to determine when the detection results can be trusted and when they cannot \cite{relatedsoftmax_fromvgg}.
    This motivates us to learn a calibrated mask score according to MaskIoU for every detection hypothesis. Without loss of generality, we work on the Mask R-CNN framework, and propose Mask Scoring R-CNN (MS R-CNN), a Mask R-CNN with an additional MaskIoU head module that learns the MaskIoU aligned mask score. The predicted mask scores of our framework are shown in Fig.~\ref{fig:comv4}~(b) and the orange histogram in Fig.~\ref{fig:comv4}~(c).

\begin{figure*}[ht]
\centering
\includegraphics[width=0.98\linewidth]{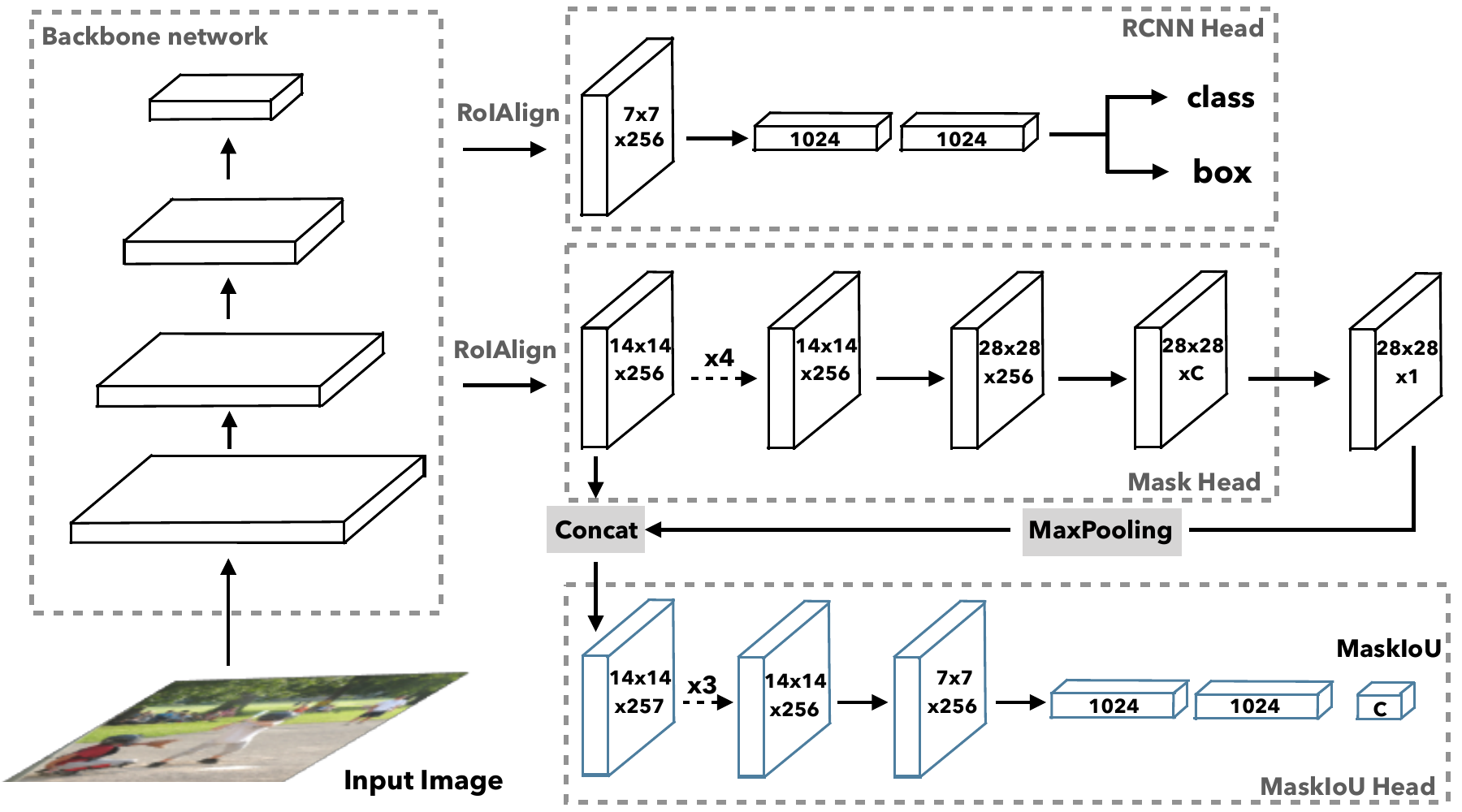}
\caption{Network architecture of Mask Scoring R-CNN. The input image is fed into a backbone network to generate RoIs via RPN and RoI features via RoIAlign. The RCNN head and Mask head are standard components of Mask R-CNN. For predicting MaskIoU, we use the predicted mask and RoI feature as input. The MaskIoU head has 4 convolution layers (all have kernel=3 and the final one uses stride=2 for downsampling) and 3 fully connected layers (the final one outputs $C$ classes MaskIoU.) }
\label{fig:network}
\end{figure*}   

%-------------------------------------------------------------------------
\subsection{Mask scoring in Mask R-CNN}
    %problem of mask rcnn
    %Current Mask R-CNN use the classification score as masks score, there will be a case that the detection with high classification score while masks are worse, as figure \ref{fig:sample} shows. For quantitative analysis, we use ResNet18 Mask R-CNN for experiments in COCO2017 validation dataset. After SoftNMS, we select the masks which IoU with ground truth masks are larger than 0.5 and the classification score is larger than 0.5. We then get the IoU and the score of each masks. As Figure \ref{fig:com} shown, Mask R-CNN will get a high score no matter what the IoU is, the low IoU masks can get a similar score with the high IoU masks, this will pull down the results. When using our method, we can see that our method can make the IoU and score more discriminative. We penalize the score of low IoU masks and retain the high score of high IoU masks. Our method using predicting IoU of masks to correct the masks score, and we can get a remarkable improvement. We will introduce our method following.
    
    Mask Scoring R-CNN is conceptually simple: Mask R-CNN with MaskIoU Head, which takes the instance feature and the predicted mask together as input, and predicts the IoU between input mask and ground truth mask, as shown in Fig.~\ref{fig:network}. We will present the details of our framework in the following sections. 
    % Mask R-CNN has three outputs for a detection hypothesis, a class label, a bounding box offset, and a mask. Mask scoring R-CNN adds a fourth branch that outputs the MaskIoU. Next, we introduce the design of the MaskIoU branch.

\vspace{-2mm}    
    \paragraph{Mask R-CNN:} We begin by briefly reviewing the Mask R-CNN \cite{he2017maskrcnn}. Following Faster R-CNN \cite{ren2015faster}, Mask R-CNN consists of two stages. The first stage is the Region Proposal Network (RPN). It proposes candidate object bounding boxes regardless of object categories. The second stage is termed as the R-CNN stage, which extracts features using RoIAlign for each proposal and performs proposal classification, bounding box regression and mask predicting.

\vspace{-2mm}  
    \paragraph{Mask scoring:} 
    % We define $s_{mask}$ as the score of the predicted mask. The ideal $s_{mask}$ is equal to the pixel-level IoU between predicted mask and its matched ground truth mask, which is termed as MaskIoU before. The dimension of $s_{mask}$ is $C\time 1$ where $C$ is the number of object categories. This indicates that for each region proposal, it has a class-specific mask score.
    % Learning the mask score has two tasks: learning the proposal classification score and and regressing proposal's MaskIoU for foreground object category. It is hard to train the two tasks only using a single objective function. For simplify, we can decompose the mask score learning task into two sub-tasks, denoted as $s_{mask} = s_{cls} \cdot s_{iou}$ for all foreground object categories. $s_{cls}$ focuses on classifying the proposal belong to which class and $s_{iou}$ focuses on regressing the MaskIoU. 
    
    We define $s_{mask}$ as the score of the predicted mask. The ideal $s_{mask}$ is equal to the pixel-level IoU between predicted mask and its matched ground truth mask, which is termed as MaskIoU before. The ideal $s_{mask}$ also should only have positive value for ground truth category, and be zero for other classes, since a mask only belong to one class. This requires the mask score to works well on two task: classifying the mask to right category and regressing the proposal's MaskIoU for foreground object category.
    
    It is hard to train the two tasks only using a single objective function. For simplify, we can decompose the mask score learning task into mask classification and IoU regression, denoted as $s_{mask} = s_{cls} \cdot s_{iou}$ for all object categories. $s_{cls}$ focuses on classifying the proposal belong to which class and $s_{iou}$ focuses on regressing the MaskIoU.

    As for $s_{cls}$, the goal of $s_{cls}$ is to classify the proposal belonging to which class, which has been done in the classification task in the R-CNN stage. So we can directly take the corresponding classification score.  
    Regressing $s_{iou}$ is the target of this paper, which is discussed in the following paragraph.

\begin{table*}[ht!]
\caption{COCO 2017 validation results. We report both detection and instance segmentation results. $AP_m$ denotes instance segmentation results and $AP_b$ denotes detection results. The results without \checkmark  are those of Mask R-CNN, while with \checkmark  are those of our MS R-CNN. The results show that our method is insensitive to different backbone networks. }
  \centering
  \vspace{3mm}
  \begin{tabular}{l|c|c c c |c c c}
  \toprule
    %& \multicolumn{3}{|c}{Avg.Precision, IoU:} & \multicolumn{3}{|c}{Avg.Precision, Area:} \\
    Backbone & MaskIoU head & $AP_{m}$ & $AP_{m}$@0.5 & $AP_{m}$@0.75 & $AP_{b}$ & $AP_{b}$@0.5 & $AP_{b}$@0.75\\
    \midrule
    \multirow{2}{*}{ResNet-18 FPN} &  & 27.7 & 46.9 & 29.0 & 31.2 & 50.4 & 33.2  \\
      & \checkmark & 29.3 & 46.9 & 31.3 & 31.5 & 50.8 &  33.5  \\
    \midrule
    \multirow{2}{*}{ResNet-50 FPN} &  & 34.5 & 55.8 & 36.7 & 38.6 & 59.2 & 42.5   \\
    &  \checkmark & 36.0 & 55.8 & 38.8 & 38.6 & 59.2 & 42.5   \\
    \midrule
    \multirow{2}{*}{ResNet-101 FPN} & & 36.6 & 58.6 & 39.0 & 41.3 & 61.7 &  45.9 \\
    &  \checkmark & 38.2 & 58.4 & 41.5 & 41.4 & 61.8 & 46.3  \\
    \bottomrule
  \end{tabular}
  \label{table:mainresultbackbone}
\end{table*}

\begin{table*}[ht!]
\caption{COCO 2017 validation results. We report detection and instance segmentation results. $AP_m$ denotes instance segmentation results and $AP_b$ denotes detection results. In the results area, rows 1\&2 use the Faster R-CNN framework; rows 3\&4 additionally use FPN framework; rows 5\&6 additionally use the DCN+FPN. The results show that consistent improvement of the proposed MaskIoU head.}
  \centering
  \vspace{3mm}
  \begin{tabular}{l|c c c | c c c | c c c}
  \toprule
    %& \multicolumn{3}{|c}{Avg.Precision, IoU:} & \multicolumn{3}{|c}{Avg.Precision, Area:} \\
    Backbone & MaskIoU head & FPN & DCN & $AP_{m}$ & $AP_{m}$@0.5 & $AP_{m}$@0.75 & $AP_{b}$ & $AP_{b}$@0.5 & $AP_{b}$@0.75  \\
    \midrule
    \multirow{6}{*}{ResNet-101} & & & & 33.9 & 53.9 & 36.2 & 38.6  &  57.3 & 42.8 \\
      & \checkmark  &  &  & 35.0 & 54.0 & 37.7 & 38.7  & 57.4  & 43.0  \\
      &  & \checkmark & & 36.6 & 58.6 & 39.0 & 41.3   & 61.7  & 45.9 \\
      & \checkmark & \checkmark & & 38.2 & 58.4 & 41.5 & 41.4  & 61.8  & 46.3  \\
      &  & \checkmark & \checkmark & 37.7 & 60.3 & 40.0 & 42.9  & 63.4  & 47.8   \\
      & \checkmark & \checkmark & \checkmark & 39.1 & 60.0 & 42.4 & 43.1  & 63.5  & 47.7 \\
    \bottomrule
  \end{tabular}
  \label{table:mainresultframework}
\end{table*} 

    \paragraph{MaskIoU head:} The MaskIoU head aims to regress the IoU between the predicted mask and its ground truth mask. We use the concatenation of feature from RoIAlign layer and the predicted mask as the input of MaskIoU head. When concatenating, we use a max pooing layer with kernel size of 2 and stride of 2 to make the predicted mask have the same spatial size with RoI feature. We only choose to regress the MaskIoU for the ground truth class (for testing, we choose the predicted class) instead of all classes. Our MaskIoU head consists of 4 convolution layers and 3 fully connected layers. For the 4 convolution layers, we follow Mask head and set the kernel size and filter number to 3 and 256 respectively for all the convolution layers. For the 3 fully connected (FC) layers, we follow the RCNN head and set the outputs of the first two FC layers to 1024 and the output of the final FC to the number of classes.

    % For predicting masks IoU, the formally type is predicting Pr(IoU$|$ $Class_i$) for each class, however, regressing masks IoU for each class may be hard. Here we decompose this task into a masks IoU classification task and masks IoU regression task: Pr($IoU_\text{class}$)*Pr(Class), predicting class and masks IoU separately. If the proposal is background, we only train Pr(Class), if the proposal is positive, we both train Pr(class) and Pr($IoU_class$), note that for 81 classes Pr($IoU_\text{class}$), we only regress the ground truth class IoU and ignore other class IoU. Masks IoU classification aims to classify the class of proposal, the goal is the same as RCNN classification, for the sake of simplicity, we can use the results of RCNN classification as the results of masks IoU classification. Then we only need to focus on masks IoU regression.
    
    %overall network
    % Our method is built on Mask R-CNN, we use ResNet as feature extractor. As shown in figure \ref{fig:network}, the network is a standard Mask R-CNN which the RCNN head is used for classification and regression and the Mask head is used for generating masks. For masks IoU regression, we only need to add a branch (called MaskIoU head).

\vspace{-2mm}
    \paragraph{Training:}

    For training the MaskIoU head, we use the RPN proposals as training samples. %  we have tried generating bounding boxes by augmenting the ground-truth instead of using RPN proposals, but it works worse than just using RPN proposals. 
    The training samples are required to have a IoU between proposal box and the matched ground truth box larger than 0.5, which are the same with the training samples of the Mask head of Mask R-CNN. For generating the regression target for each training sample, we firstly get the predicted mask of the target class and binarize the predicted mask using a threshold of 0.5 
    %\xw{Binarization is not shown in Fig.~\ref{fig:fusiontypes}}(这是生成训练label的，和怎么fusing没关系). 

    Then we use the MaskIoU between the binary mask and its matched ground truth as the MaskIoU target.
    We use the $\ell_2$ loss for regressing MaskIoU, and the loss weight is set to $1$. The proposed MaskIoU head is integrated into Mask R-CNN, and the whole network is end to end trained.

\vspace{-2mm}
    \paragraph{Inference:}
    
    % The R-CNN stage of Mask R-CNN outputs $N$ bounding boxes; among them top-$k$ (\ie $k=100$) scoring boxes after SoftNMS \cite{bodla2017softnms} are selected; then the top-$k$ boxes are fed into the Mask head to generate multi-class masks. Our method follows this procedure, we only feed the top-$k$ target masks to predict the MaskIoU, and multiply the classification score by the predicted MaskIoU; finally the new score is used as the masks score.
    
    During inference, we just use MaskIoU head to calibrate classification score generated from R-CNN. Specifically, suppose the R-CNN stage of Mask R-CNN outputs $N$ bounding boxes, and among them top-$k$ (\ie $k=100$) scoring boxes after SoftNMS \cite{bodla2017softnms} are selected. Then the top-$k$ boxes are fed into the Mask head to generate multi-class masks. This is the standard Mask R-CNN inference procedure. We follow this procedure as well, and feed the top-$k$ target masks to predict the MaskIoU. The predicted MaskIoU are multiplied with classification score, to get the new calibrated mask score as the final mask confidence.

\begin{table*}[ht!]
\caption{Comparing different instance segmentation methods on COCO 2017 test-dev.}
  \centering
  \vspace{3mm}
  \begin{tabular}{l|c|c c c|c c c}
  \toprule
    %& \multicolumn{3}{|c}{Avg.Precision, IoU:} & \multicolumn{3}{|c}{Avg.Precision, Area:} \\
    Method & Backbone & AP & AP@0.5 & AP@0.75 & AP${_S}$ & AP${_M}$ & AP${_L}$\\
    \midrule
    MNC \cite{dai2016mnc} &  ResNet-101 & 24.6 & 44.3 & 24.8 & 4.7 & 25.9 & 43.6\\
    FCIS \cite{fcis} &  ResNet-101 & 29.2 & 49.5 & - & - & - & -\\
    FCIS+++ \cite{fcis}  & ResNet-101 & 33.6 & 54.5 & - & - & - & -\\
    Mask R-CNN \cite{he2017maskrcnn} & ResNet-101 & 33.1 & 54.9 & 34.8 & 12.1 & 35.6 & 51.1\\
    Mask R-CNN \cite{he2017maskrcnn}  & ResNet-101 FPN & 35.7 & 58.0 & 37.8 & 15.5 & 38.1 & 52.4\\ 
    Mask R-CNN \cite{he2017maskrcnn}&  ResNeXt-101 FPN & 37.1 & 60.0 & 39.4 & 16.9 & 39.9 & 53.5\\
    MaskLab \cite{chen2017masklab}  & ResNet-101 & 35.4 & 57.4 & 37.4 & 16.9 & 38.3 & 49.2\\
    MaskLab+ \cite{chen2017masklab} & ResNet-101 & 37.3 & 59.8 & 36.6 & 19.1 & 40.5 & 50.6\\
    MaskLab+ \cite{chen2017masklab}  & ResNet-101 (JET) & 38.1 & 61.1 & 40.4 & 19.6 & 41.6 & 51.4\\ 
    %\midrule
    %Mask R-CNN &  \multirow{2}{*}{ResNet-50}  & 31.9 & 51.7 & 34.0 & 11.9 & 33.5 & 49.3\\
    %MS R-CNN  &    & 33.1 & 51.8 & 35.7 & 12.5 & 34.7 & 50.7\\
    \midrule
    Mask R-CNN & \multirow{2}{*}{ResNet-101}  & 34.3 & 55.0 & 36.6 & 13.2 & 36.4 & 52.2\\
    MS R-CNN  &    & 35.4 & 54.9 & 38.1 & 13.7 & 37.6 & 53.3\\ 
    %\midrule
    %Mask R-CNN & \multirow{2}{*}{ResNet-50 FPN} & 34.8 & 56.5 & 37.1 & 16.2 & 36.6 & 49.7\\
    %MS R-CNN  &   & 36.2 & 56.1 & 39.3 & 16.6 & 37.7 & 51.4\\
    \midrule
    Mask R-CNN & \multirow{2}{*}{ResNet-101 FPN} & 37.0 & 59.2 & 39.5 & 17.1 & 39.3 & 52.9\\
    MS R-CNN  &   & 38.3 & 58.8 & 41.5 & 17.8 & 40.4 & 54.4\\
    %\midrule
    %Mask R-CNN & \multirow{2}{*}{ResNet-50 DCN+FPN} & 37.1 & 59.5 & 39.7 & 17.3 & 38.9 & 53.5\\
    %MS R-CNN  &   & 38.4 & 59.1 & 41.6 & 18.3 & 39.8 & 54.6\\
    \midrule
    Mask R-CNN & \multirow{2}{*}{ResNet-101 DCN+FPN}  & 38.4 & 61.2 & 41.2 & 18.0 & 40.5 & 55.2\\
    MS R-CNN  &    & 39.6 & 60.7 & 43.1 & 18.8 & 41.5 & 56.2\\
    \bottomrule
  \end{tabular}
  \label{table:testdevresults}
\end{table*}

\section{Experiments}

    All experiments are conducted on the COCO dataset \cite{lin2014microsoftcoco} with 80 object categories. We follow COCO 2017 settings, using the 115k images \emph{train} split for training, 5k \emph{validation} split for validation, 20k \emph{test-dev} split for test. We use COCO evaluation metrics AP (averaged over IoU thresholds) to report the results, including AP@0.5, AP@0.75, and $AP_{S}$, $AP_{M}$, $AP_{L}$ (AP at different scales). 
    AP@0.5 (or AP@0.75) means using an IoU threshold 0.5 (or 0.75) to identify whether a predicted bounding box or mask is positive in the evaluation.
    Unless noted, AP is evaluated using mask IoU.
    
\subsection{Implementation Details}
    We use our reproduced Mask R-CNN for all experiments. We use ResNet-18 based FPN network for ablation study and ResNet-18/50/101 based on Faster R-CNN/FPN/DCN+FPN \cite{dai2017deformable} for comparing our method with other baseline results. For ResNet-18 FPN, input images are resized to have 600px along the short axis and a maximum of 1000px along the long axis for training and testing. Different from the standard FPN \cite{lin2017fpn}, we only use C4, C5 for RPN proposal and feature extractor in ResNet-18. For ResNet-50/101, input images are resized to 800 px for the short axis and 1333px for the long axis for training and testing. The rest configurations for ResNet-50/101 follow Detectron \cite{Detectron2018}. 
    We train all the networks for 18 epochs, decreasing the learning rate by a factor of 0.1 after 14 epochs and 17 epochs. Synchronized SGD with momentum 0.9 is used as optimizer. For testing, we use SoftNMS and retain the top-100 score detection for each image.
    %and then after the Mask head and MaskIoU head, we can get the predicted MaskIoU. The predicted MaskIoU are multiplied with classification score, getting the new calibrated mask score as the final mask score.

\subsection{Quantitative Results}

    We report our results on different backbone networks including ResNet-18/50/101 and different framework including Faster R-CNN/FPN/DCN+FPN \cite{dai2017deformable} to prove the effectiveness of our method. Results are shown in Table~\ref{table:mainresultbackbone} and Table~\ref{table:mainresultframework}. We use $AP_m$ to report instance segmentation results and $AP_b$ to report detection results. We report our reproduced Mask R-CNN results and our MS R-CNN results. As Table~\ref{table:mainresultbackbone} shows, comparing with Mask R-CNN, our MS R-CNN is not sensitive to the backbone network and can achieve stable improvement on all backbone networks: Our MS R-CNN can get a remarkable improvement (about 1.5 AP). Especially for AP@0.75, our method can improve baseline by about 2 points.  Table~\ref{table:mainresultframework} indicates that our MS R-CNN is robust to different framework including Faster R-CNN/FPN/DCN+FPN. Beside, our MS R-CNN does not harm bounding box detection performance; in fact, it improves bounding box detection performance slightly. The results of \emph{test-dev} are reported in Table~\ref{table:testdevresults}, only the instance segmentation results are reported. 
    %We compare to the recent instance segmentation methods, such as MaskLab \cite{chen2017masklab}, Mask R-CNN \cite{he2017maskrcnn} and FCIS \cite{fcis}, our MS R-CNN obtains the state-of-the-art instance segmentation performance.
    %我们复现的mask rcnn已经比他们的高了，感觉这样说不是很好吧。。。

\subsection{Ablation Study}
    We comprehensively evaluate our method on COCO 2017 validation set. We use ResNet-18 FPN for all the ablation study experiments.
    
%\subsubsection{The design choices of MaskIoU head}
\paragraph{The design choices of MaskIoU head input:}

\begin{figure*}
\centering
\includegraphics[width=0.9\linewidth]{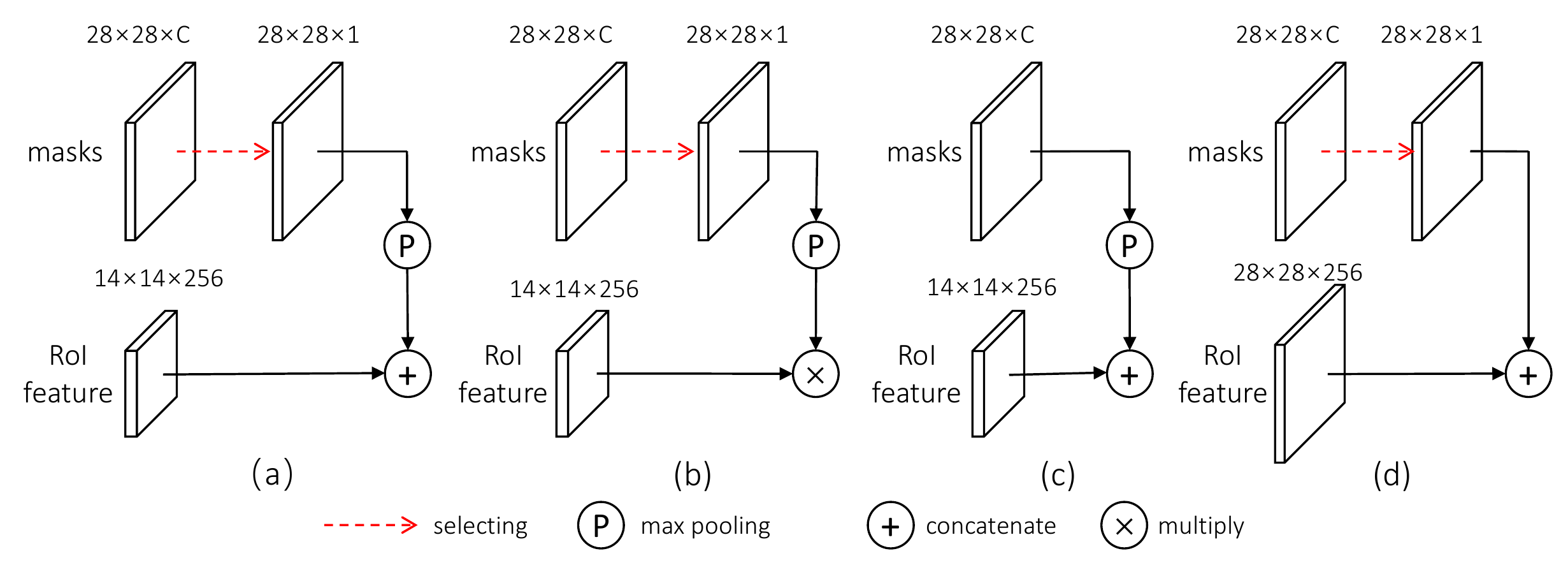}
\caption{Different design choices of the MaskIoU head input.}  
\label{fig:fusiontypes}
\end{figure*} 

    % We first study the design choices of the MaskIoU head \lichao{input,}. This head takes the mask score map ($28\times\!28\times\!81$) predicted from the mask head and the RoI features as inputs,  and outputs a feature map for MaskIoU prediction. There are a few design choices shown in Fig.~\ref{fig:fusiontypes} and explained as follows:
    
    We first study the design choices of the MaskIoU head input, which is the fusion of predicted mask score map ($28\times\!28\times\!C$) from the mask head and the RoI features. There are a few design choices shown in Fig.~\ref{fig:fusiontypes} and explained as follows:

    \begin{itemize}
    \setlength\itemsep{0.0em}
    \item[(a)] Target mask concatenates RoI feature: The score map of the target class is taken, max-pooled and concatenated with RoI feature.
    \item[(b)] Target mask multiplies RoI feature: The score map of the target class is taken, max-pooled and multiplied with RoI feature.
    \item[(c)] All masks concatenates RoI feature: All the C classes mask score map are max-pooled and concatenated with RoI feature.
    \item[(d)] Target mask concatenates High-resolution RoI feature: The score map of the target class is taken and concatenated with $28\times\!28$ RoI features.
    \end{itemize}

    % The results are shown in Table~\ref{table:fusingtypes}. The results show that the performance of MaskIoU head is robust to different design choices and all of them can obviously outperform the Mask R-CNN baseline. Since concatenating the target score map and RoI feature obtains the best results, we use it as our default choice. 

    The results are shown in Table~\ref{table:fusingtypes}. We can see that the performance of MaskIoU head is robust to different ways of fusing mask prediction and RoI feature. Performance gain is observed in all kinds of design. Since concatenating the target score map and RoI feature obtains the best results, we use it as our default choice. 

\begin{table}
\caption{Results of different design choices of the MaskIoU head input.}
\vspace{3mm}
  \centering
  %\resizebox{\textwidth}{!}
  {\begin{tabular}{l|c c c}
  \toprule
    Setting & AP & AP@0.5 & AP@0.75 \\
    \midrule
    Mask R-CNN baseline & 27.7 & 46.9 & 29.0 \\
    (a) Target mask $+$ RoI & 29.3 & 46.9 & 31.3 \\
    (b) Target mask $\times$ RoI & 29.1 & 46.6 & 30.9 \\
    (c) All masks $+$ RoI & 29.1 & 46.6 & 30.8 \\
    (d) Target mask $+$ HR RoI & 29.1 & 46.7 & 31.1 \\
    \bottomrule
  \end{tabular}}
  \label{table:fusingtypes}
\end{table}

%\subsubsection{The choices of the training target}
\paragraph{The choices of the training target:}
    % As mentioned above, a proposal usually contains multiple objects. Thus, setting the training target of MaskIoU head involves multiple objects. There are different choices are listed as follows:

    As mentioned before, we decompose the mask score learning task as mask classification and MaskIoU regression. Is it possible to learn the mask score directly? 
    %In addition, a predicted mask may belong to multiple categories due to the imperfect mask segmentation. 
    In addition, a RoI may contain multiple categories of objects. Should we learn MaskIoU for all categories? How to set the training target for MaskIoU head still need exploration. There are many different choices of training target:

    \begin{enumerate}
    \setlength\itemsep{0.0em}
        \item Learning the MaskIoU of the target category, meanwhile the other categories in the proposal are ignored. This is also the default training target in this paper, and the control group for all experiments in this paragraph.
        \item Learning the MaskIoU for all categories. If a category does not appear in the RoI, its target MaskIoU is set to 0. This setting denotes using regression only to predict MaskIoU, which requires the regressor to be aware of the absence of unrelated categories. 
        \item Learning the MaskIoU of all the positive categories, where a positive category means the category appears in the RoI region. And the rest categories in the proposal are ignored. This setting is used to see whether perform regression for more categories in the RoI region could be better.
    \end{enumerate}

    % \begin{enumerate}
    % \setlength\itemsep{0.0em}
    %     \item Learning the MaskIoU of the target instance, meanwhile the rest instances and the unseen class in the proposal are ignored.
    %     \item Learning the MaskIoU of all the positive instances, where a positive instance means it appears in the RoI region, meanwhile the rest instances and the unseen class in the proposal are ignored.
    %     \item Learning the MaskIoU for all categories. If a category does not appear in the RoI , its target MaskIoU is set to 0.
    %     \item Learning the MaskIoU of all the positive instances using all the score masks as features, as shown in Fig.~\ref{fig:fusiontypes}~(c). Note that the above setting \#2 and \#3 both only use the target mask as feature. 
    %     \item Learning the MaskIoU of all categories, as the above setting \#3, using all the score masks as features.
    % \end{enumerate}

\begin{table}[ht!]
\caption{Results of using different training targets.}
\vspace{3mm}
  \centering
  %\resizebox{\textwidth}{!}
  {\begin{tabular}{l|c c c}
  \toprule
    Setting & AP & AP@0.5 & AP@0.75 \\
    \midrule
    Mask R-CNN baseline & 27.7 & 46.9 & 29.0 \\
    Setting \#1: Target ins. & 29.3 & 46.9 & 31.3 \\
    Setting \#2: All cls.  & 24.5 & 41.6 & 25.6 \\
    Setting \#3: Positive ins.  & 28.2 & 45.5 & 30.2 \\
    % Positive ins. + all masks  & 28.2 & 45.5 & 30.1 \\
    % All cls. + all masks  & 25.1 & 42.1 & 26.2 \\
    \bottomrule
  \end{tabular}}
  \label{table:trainallclasses}
\end{table}  

    % Table~\ref{table:trainallclasses} shows the results for above training targets. By comparing setting \#1 with setting \#2, we can find that training MaskIoU of all categories will pull down the experimental results. This is because the MaskIoU to be trained needs to classify the category of the object and regress MaskIoU of this category, which includes classification and regression tasks, that verifies our thought that training classification and regression using a single objective function will be difficult. 
    % Although learning MaskIoU for all positive instances can outperform Mask R-CNN baseline, it is still worse than only learning for the target instance. 
    
    Table~\ref{table:trainallclasses} shows the results for above training targets. By comparing setting \#1 with setting \#2, we can find that training MaskIoU of all categories (regression only based MaskIoU prediction) will degrade the performance drastically, which verifies our opinion that training classification and regression using a single objective function is difficult.

    It is reasonable that the performance of setting \#3 is inferior to setting \#1, since regressing MaskIoU for all positive categories increases the burden of MaskIoU head.
    Thus, learning the MaskIoU of the target category is used as our default choice.
    %while only the categories with largest classification score is kept for mask score calibration.

%\subsubsection{How to select training samples}
\paragraph{How to select training samples:}

    Since the proposed MaskIoU head is built on top of the Mask R-CNN framework, all the training samples for the MaskIoU head have a box-level IoU larger than 0.5 with its ground truth bounding box according to the setting in the Mask R-CNN. However, their MaskIoU may not exceed 0.5. 
    
    Given a threshold $\tau$, we use the samples whose MaskIoU are larger than $\tau$ to train the MaskIoU head. Table~\ref{table:trainingsamples} shows the results. The results show that training using all the examples obtains the best performance.

\begin{table}
\caption{Results of selecting different training samples for the MaskIoU head.}
\vspace{3mm}
  \centering
  %\resizebox{\textwidth}{!}
  \setlength{\tabcolsep}{3mm}
  {\begin{tabular}{c|c c c}
  \toprule
    Threshold & AP & AP@0.5 & AP@0.75 \\
    \midrule
    $\tau = 0.0$ & 29.3 & 46.9 & 31.3 \\
    $\tau = 0.3$  & 29.2 & 46.6 & 31.1 \\
    $\tau = 0.5$  & 29.0 & 46.5 & 30.9 \\
    $\tau = 0.7$  & 28.8 & 46.9 & 30.5 \\
    \bottomrule
  \end{tabular}}
  \label{table:trainingsamples}
\end{table}

\subsection{Discussion}

    In this section, we will first discuss the quality of the predicted MaskIoU, and then investigate the upper bound performance of Mask Scoring R-CNN if the prediction of MaskIoU is perfect, and analyze the computational complexity of MaskIoU head at last.
    In the discussions, all the results are obtained on COCO 2017 validation set using both a weak backbone network, \ie, ResNet-18 FPN and a strong backbone network, \ie, ResNet-101 DCN+FPN.
    
%\subsubsection{The quality of the predicted MaskIoU}
\paragraph{The quality of the predicted MaskIoU:}
    We use correlation coefficient between ground truth and predicted MaskIoU to measure the quality of our prediction. Reviewing our testing procedure, we choose the top 100 scoring boxes after SoftNMS according to the classification scores, fed the detected boxes to Mask head and get the predicted mask, then use the predicted mask and RoI feature as the input of MaskIoU head. The output of MaskIoU head and classification score are further integrated into final mask score. 
    % Since the COCO 2017 validation set contains $5,000$ images and every has $100$ predicted MaskIoU, there are in total $500,000$ predictions. The predictions and their corresponding ground truth MaskIoU are visualized in Fig.~\ref{fig:scatter}. 
    
\begin{figure}[!t]
\centering
\includegraphics[width=1.0\linewidth]{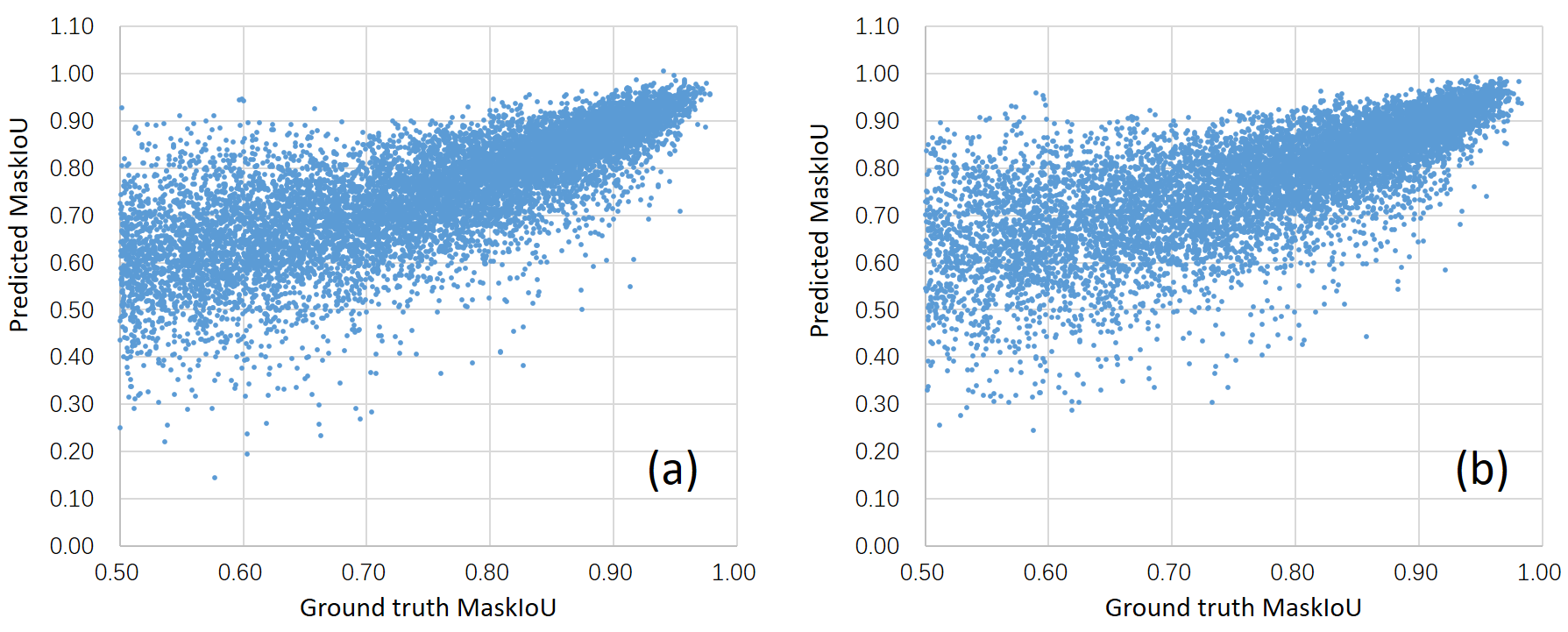}
\caption{Visualizations of MaskIoU predictions and their ground truth. (a) Results with ResNet-18 FPN backbone and (b) results with ResNet-101 DCN+FPN backbone. The x-axis presents the ground truth MaskIoU and the y-axis presents the predicted MaskIoU of the proposed MaskIoU head.}
\label{fig:scatter}
\end{figure}

    We keep $100$ predicted MaskIoU for each image in the COCO 2017 validation dataset, collecting $500,000$ predictions from all $5,000$ images. We plot each predictions and their corresponding ground truth in Fig.~\ref{fig:scatter}.
    % We can see that the MaskIoU predictions have good correlation with their ground truth, especially, the predictions with high MaskIoU obtain better correlation.
    We can see that the MaskIoU predictions have good correlation with their ground truth, especially for those prediction with high MaskIoU.
    % For quantitative analysis, we calculate the correlation coefficient between predictions and their ground truth. No matter using ResNet-18 FPN as backbone or using ResNet-101 DCN+FPN as backbone, the correlation coefficients are both  \textbf{$0.74$}.
    % Comparing the results of using different backbones, we find that the quality of the prediction is not sensitive to the change of backbone.
    The correlation coefficient between predictions and their ground truth is around \textbf{$0.74$} for both ResNet-18 FPN and ResNet-101 DCN+FPN backbone networks. It indicates that the quality of the prediction is not sensitive to the change of backbone networks. This conclusion is also consistent with Table~\ref{table:mainresultbackbone}. 
    Since there is no method works on predicting MaskIoU before, we refer to a previous work \cite{jiang2018acquisition} on predicting bounding box IoU.  \cite{jiang2018acquisition} obtains a $0.617$ correlation coefficient, which is inferior to ours.

%\subsubsection{The upper bound performance of Mask scoring R-CNN}
\paragraph{The upper bound performance of MS R-CNN:}

    Here we will discuss the upper bound performance of our method. For each predicted mask, we can find its matched ground truth mask; then we just use the ground truth MaskIoU to replace the predicted MaskIoU when the ground truth MaskIoU larger than 0. The results are shown in Table~\ref{table:upper_limit}. 
    The results show that Mask Scoring R-CNN consistently outperforms Mask R-CNN. %compared to the ideal performance of Mask scoring R-CNN, there is still a room to improve the practical Mask scoring R-CNN, which are $2.2\%$ mAP for ResNet-18 FPN backbone and $2.6\%$ mAP for ResNet-101 DCN+FPN backbone.
    Compared to the ideal prediction of Mask Scoring R-CNN, there is still a room to improve the practical Mask Scoring R-CNN, which are $2.2\%$ AP for ResNet-18 FPN backbone and $2.6\%$ AP for ResNet-101 DCN+FPN backbone.
    
    % To further improve the performance of Mask scoring R-CNN, we may refer to Fig.~\ref{fig:scatter}; what we need to improve is the prediction accuracies of the ``bad" masks, \ie, the masks that have low MaskIoU with their ground truth.
    
    % MaskIoU 低的,和 groundtruth差距大， 原因还可能是本身classfication 不准。因为只有classification 值为0 或1的二值，才符合模型的假设。所以不提上面的

\begin{comment}
\begin{table}[ht!]
\caption{Upper limit results. We use ResNet-18 for experiments in COCO 2017 validation dataset. use gt IoU(\emph{thresh}) means if the ground truth MaskIoU is higher than \emph{thresh}, then use the ground truth MaskIoU to replace the predicting MaskIoU, otherwise, still using predicting MaskIoU. }
\vspace{3mm}
  \centering
  %\resizebox{\textwidth}{!}
  \setlength{\tabcolsep}{3mm}
  {\begin{tabular}{c|c c c}
  \toprule
    Method & AP & AP@0.5 & AP@0.75 \\
    \midrule
    baseline & 27.7 & 46.9 & 29.0 \\
    predicted MaskIoU & 29.3 & 46.9 & 31.3 \\
    gt MaskIoU   & 31.5 & 48.4 & 34.2 \\
    \midrule
    baseline & 37.7 & 60.3 & 40.0 \\
    predicted MaskIoU & 39.1 & 60.0 & 42.4 \\
    gt MaskIoU  & 41.7 & 61.5 & 45.7 \\
    \bottomrule
  \end{tabular}}
  \label{table:upper_limit}
\end{table}
\end{comment}

\begin{table}
\caption{Results of Mask R-CNN, MS R-CNN and the ideal case of MS R-CNN (MS R-CNN$^\star$) using ResNet-18 FPN and ResNet-101 DCN+FPN as backbones on COCO 2017 validation set.}
\vspace{3mm}
  \centering
  %\resizebox{\textwidth}{!}
  \setlength{\tabcolsep}{3mm}
  {\begin{tabular}{c|c|c}
  \toprule
    Method & Backbone & AP\\
    \midrule
    Mask R-CNN & \multirow{3}{*}{ResNet-18 FPN} & 27.7  \\
    MS R-CNN && 29.3 \\
    MS R-CNN$^\star$&   & 31.5  \\
    \midrule
    Mask R-CNN & \multirow{3}{*}{ResNet-101 DCN+FPN} & 37.7  \\
    MS R-CNN & & 39.1 \\
    MS R-CNN$^\star$ &  & 41.7  \\
    \bottomrule
  \end{tabular}}
  \label{table:upper_limit}
\end{table}

%\subsubsection{Complexity}
\paragraph{Model size and running time:}
    Our MaskIoU head has about 0.39G FLOPs while Mask head has about 0.53G FLOPs for each proposal. 
    %As Fig.~\ref{fig:network} shows, no matter what backbone network we use, the FLOPs of MaskIoU head are the same. 
    We use one TITAN V GPU to test the speed (sec./image). As for ResNet-18 FPN, the speed is about 0.132 for both Mask R-CNN and MS R-CNN. As for ResNet-101 DCN+FPN, the speed is about 0.202 for both Mask R-CNN and MS R-CNN. The computation cost of MaskIoU head in Mask Scoring R-CNN is negligible. 

\section{Conclusion}
    % In this paper, we investigate the problem of scoring instance segmentation masks and propose Mask scoring R-CNN. By adding a MaskIoU head in Mask R-CNN, scores of the masks which have a high classification score while do not have good IoU with their corresponding target masks are penalized. 
    % The proposed MaskIoU head is extremely effective and easily to implement. On the COCO benchmark, extensive results show that Mask Scoring R-CNN consistently and obviously outperform Mask R-CNN. It also can be applied in other instance segmentation networks to obtain more reliable mask scores.
    In this paper, we investigate the problem of scoring instance segmentation masks and propose Mask Scoring R-CNN. By adding a MaskIoU head in Mask R-CNN, scores of the masks are aligned with MaskIoU, which is usually ignored in most instance segmentation frameworks. 
    The proposed MaskIoU head is extremely effective and easy to implement. On the COCO benchmark, extensive results show that Mask Scoring R-CNN consistently and obviously outperforms Mask R-CNN. It also can be applied to other instance segmentation networks to obtain more reliable mask scores. We hope our simple and effective approach will serve as a baseline and help the future research in instance segmentation task.

{\small
\bibliographystyle{ieee}
\bibliography{egbib}
}

\end{document}